%% file: main.tex
\documentclass[letterpaper, 10 pt, conference]{ieeeconf}  

\IEEEoverridecommandlockouts                              

\usepackage{hyperref}
\usepackage{booktabs}
\usepackage{multirow}
\usepackage{float}
\usepackage[hypcap=false]{caption}
\usepackage{graphicx}
\usepackage[section]{placeins}
\usepackage{etoolbox,lipsum}
\usepackage{afterpage}
\usepackage{booktabs}
\usepackage{subcaption}
\usepackage{capt-of,etoolbox}
\usepackage[section]{placeins}
\usepackage{etoolbox,lipsum}
\usepackage{multirow, makecell}
\usepackage{multirow}
\usepackage{gensymb}
\usepackage[usenames,dvipsnames]{xcolor}
\usepackage{soul}
\usepackage{color}
\newcommand{\mypar}[1]{\vspace{0.1cm}\noindent\textbf{#1}.}

\usepackage{expl3}
\usepackage{svg}
\ExplSyntaxOn
\newcommand\latinabbrev[1]{
	\peek_meaning:NTF . {
		#1\@}%
	{ \peek_catcode:NTF a {
			#1.\@ }%
		{#1.\@}}}
\ExplSyntaxOff
\def\eg{\latinabbrev{e.g}}
\def\etal{\latinabbrev{et al}}

\usepackage{tikz}

\usepackage{balance}
\usepackage{stfloats}

\usepackage{hyperref}
\hypersetup{
    colorlinks=true,
    linkcolor=black,
    filecolor=black,      
    urlcolor=magenta,
    pdftitle={Overleaf Example},
    pdfpagemode=FullScreen,
    }
\definecolor{lightsteelblue}{RGB}{176,196,222}
\definecolor{lightsteelred}{RGB}{230,176,160}
\definecolor{lightsteellila}{RGB}{175,181,224}
\definecolor{lightsteelgreen}{RGB}{182,214,207}
\definecolor{lightsteelyellow}{RGB}{240,240,160}
\definecolor{lightsteelwhite}{RGB}{255,255,255}
\definecolor{lightsteelgray}{RGB}{205,201,201}
\definecolor{lightsteellightgray}{RGB}{210,210,210}

\definecolor{pp_blue}{RGB}{68,114,196}
\definecolor{pp_orange}{RGB}{237,125,49}
\definecolor{pp_lila}{RGB}{112,48,160}
\definecolor{pp_ygreen}{RGB}{112,173,71}

\definecolor{gg_blue}{RGB}{52,138,189}
\definecolor{gg_red}{RGB}{226,74,51}
\definecolor{gg_lila}{RGB}{152,142,213}
\definecolor{gg_green}{RGB}{112,173,71}
\definecolor{under70}{RGB}{255, 255, 255}
\definecolor{overone}{RGB}{192,196,192}
\definecolor{table_standard}{RGB}{230,153,0}
\definecolor{table_uncertainty}{RGB}{112,48,160}

\usepackage{tikz}
\usepackage{pifont} 
\usepackage[cp1251]{inputenc}
\def\endthebibliography{%
	\def\@noitemerr{\@latex@warning{Empty `thebibliography' environment}}%
	\endlist
}

\usepackage{lipsum}

\usepackage{tikz}

\newcommand\copyrighttext{%
	\footnotesize Accepted at IROS 2021, © IEEE. Personal use is permitted, but republication/redistribution requires IEEE permission.  Permission from IEEE must be obtained for all other uses, in any current or future media,including reprinting/republishing this material for advertising or promotional purposes, creating new collective works, for resale or redistribution toservers or lists, or reuse of any copyrighted component of this work in other works.}

\newcommand\copyrightnotice{%
	\begin{tikzpicture}[remember picture,overlay]
	\node[anchor=south,yshift=10pt] at (current page.south) {\fbox{\parbox{\dimexpr\textwidth-\fboxsep-\fboxrule\relax}{\copyrighttext}}};
	\end{tikzpicture}%
}

\title{\LARGE \bf \textit{Let's Play for Action}: Recognizing Activities of Daily Living by \\Learning from Life Simulation Video Games}

\author{Alina Roitberg$^\star$ \quad \quad \quad \quad David Schneider$^\star$  	\\
	\\
	\hspace{-0.5cm} 
	Aulia Djamal \quad \quad  Constantin Seibold \quad \quad 

  {Simon Rei\ss }  \quad \quad  Rainer Stiefelhagen
	\\
	\\Institute for Anthropomatics and Robotics
	\\ Karlsruhe Institute of Technology, Germany
	\\  {\tt\small \{firstname.lastname\}@kit.edu}%
	\\

\thanks{*David Schneider and Alina Roitberg contributed equally to this work.}

}

\begin{document}

\maketitle
\copyrightnotice
\thispagestyle{empty}
\pagestyle{empty}

\input{sections/abstract}

\input{sections/introduction}

\input{sections/related_work}

\input{sections/dataset}

\input{sections/approach}

\input{sections/evaluation}

\input{sections/conclusion}

{\small
	
	\bibliographystyle{IEEEtran}
	\bibliography{egbib}
}

\end{document}

%% file: sections/abstract.tex

\begin{abstract}

Recognizing Activities of Daily Living (ADL) is a vital process for intelligent assistive robots, but collecting large annotated datasets requires time-consuming temporal labeling and raises privacy concerns, \eg, if the data is collected in a real household.
In  this  work,  we  explore  the  concept  of constructing training examples for ADL recognition by playing life simulation video games and  introduce  the  \textsc{Sims4Action}  dataset created  with the  popular  commercial  game \textsc{the Sims 4}. 
We build \textsc{Sims4Action} by  specifically  executing  actions-of-interest  in  a ``top-down"  manner,  while  the  gaming  circumstances  allow  us  to freely switch between environments, camera angles and subject appearances.
While ADL recognition on gaming data is interesting from the theoretical perspective, the key challenge arises from transferring  it  to  the  real-world  applications,  such  as  smart-homes or assistive robotics. To meet this requirement, \textsc{Sims4Action} is accompanied with a \textsc{Gaming$\rightarrow$Real} benchmark, where the models are evaluated on real videos derived from an existing ADL dataset. 
We integrate two modern algorithms for video-based activity recognition in our framework, revealing the value of life simulation video games as an inexpensive and far less intrusive source of training data.
However, our results also indicate that tasks involving a mixture of gaming and real data are challenging, opening a new research direction.
We will make our dataset publicly available at \url{https://github.com/aroitberg/sims4action}.
\end{abstract}

%% file: sections/introduction.tex

\section{Introduction}
\begin{figure}\centering
	\includegraphics[width=\linewidth]{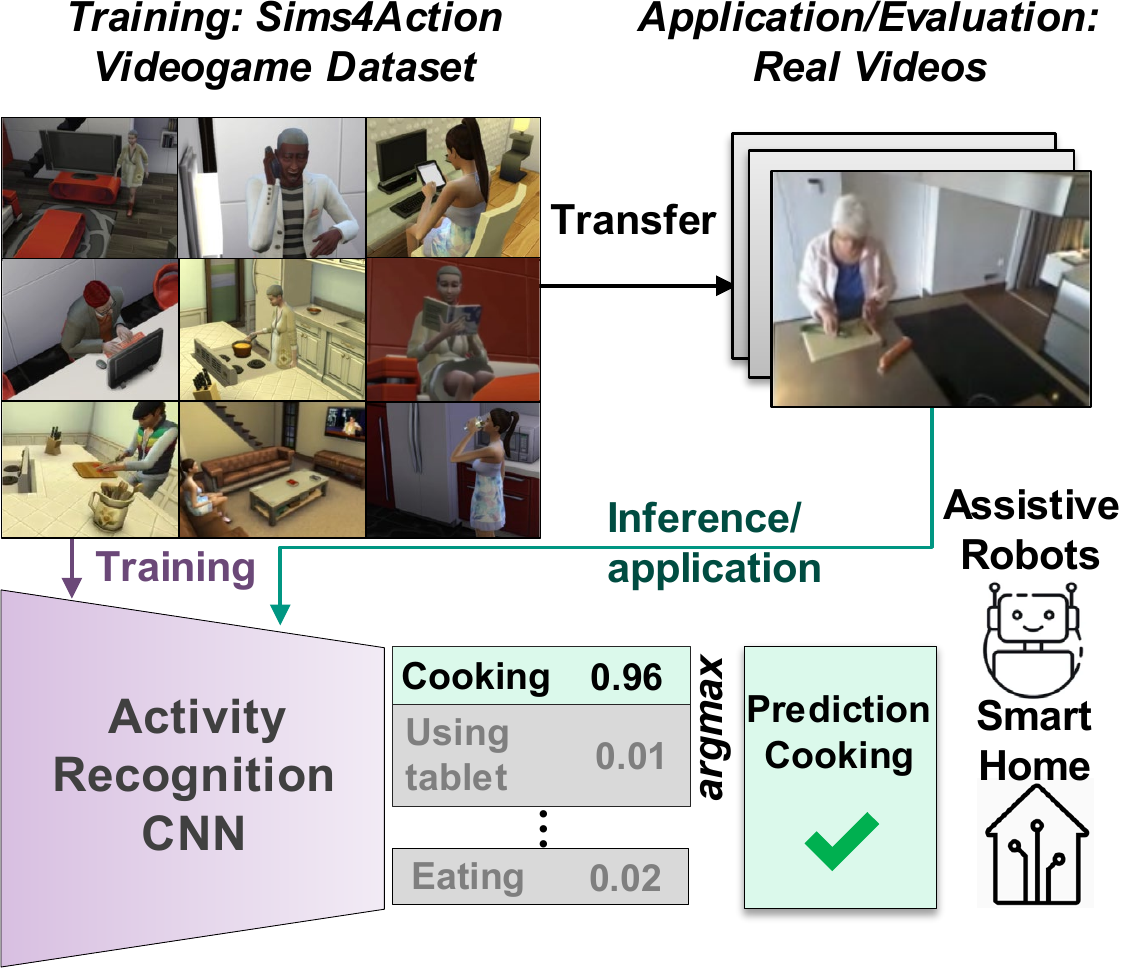}
	\caption{We leverage the life simulation video game \textsc{the Sims 4} for inexpensive ADL training data which we then utilize in the  \textsc{Gaming$\rightarrow$Real} setting in order to suit real-world applications, such as smart homes or asssistive robots. }
	\label{fig:overview_paper}
\end{figure}

With growing elderly population\footnote{Demography - Elderly population - OECD Data: \href{data.oecd.org/pop/elderly-population.htm}{\color{black}{data.oecd.org/pop/elderly-population.htm}}}, assistive household robots increasingly gain attention, but  such  technologies also  require  raising  the  levels  of  robot's  awareness.
Understanding human behaviour enables better assistance, improves human-robot-communication, situation-dependent planning and safety~\cite{christoforou2019overview,leo2018deep, marco2018computer}.
This perception task is often referred to as recognizing Activities  of  Daily  Living (ADL) and is addressed by training deep neural networks on carefully curated and manually labelled  datasets~\cite{shahroudy2016ntu,Sigurdsson2016HollywoodIH,jang2020etri,das2020vpn,das2019toyota}.

The quality of recognition is directly linked to the amount of labelled examples available during training~\cite{sun2017revisiting}.
However, collecting large video datasets tailored for a specific use-case is expensive and requires application-dependent sensory setups, recruitment of diverse participants and time-consuming accurate temporal annotation of activities.
Even after this effort is put in, the collected datasets often become insufficient, since the environment or the desired behaviour types to be recognized  may change and the costly process has to be revisited.
On top of this, data protection and privacy concerns make recruitment of participants for recording training videos  in real domestic environments even more difficult, as seen, \eg, by the blurred faces in the Toyota Smarthome dataset~\cite{das2019toyota} for ADL recognition.

Video games constitute a multi-billion dollar industry{\footnote{ Statistics on the videogame market obtained from  { \href{www.statista.com/topics/868/video-games/}{ www.statista.com/topics/868/video-games/}}}}, where  developers put great effort into build highly realistic worlds.
Intuitively we connect gameplay exclusively with entertainment, but the impressive amount of details and diversity of objects, environments, appearances, human body motion and character traits coupled with the power to control the virtual world render it an excellent source of training data.
While adventure-driving games, such as  Grand Theft Auto, have been recently leveraged as a data source for semantic segmentation in autonomous vehicles applications~\cite{richter2016playing,richter2017playing}, the potential of life simulation games for robot learning in assisted living scenarios has been left untouched and is the main motivation of our work.

We view life simulation video games as an  opportunity for inexpensive collection of visual training examples for smart homes and domestic robots and introduce the \textsc{Sims4Action} dataset built with the popular commercial game  \textsc{The Sims 4} \footnote{The Sims 4 is a commercial game by Electronic Arts (EA): \href{ea.com/en-gb/games/the-sims/the-sims-4}{ea.com/en-gb/games/the-sims/the-sims-4}}.
Our dataset features ten hours of virtual subjects engaged in different household activities, such as cooking or working on a computer. 
As \textsc{The Sims 4} is specifically designed to give the player full control of everyday situations, we are able to freely design the environments (six different locations), camera angles (four fixed cameras per location and a moving camera mode) and human appearances (four elderly people, two adults and two young adults and equal distribution of males and females).
We capture ten different behaviour categories, which we have triggered by ``playing" the desired ADL, and accompany our dataset with the \textsc{Gaming$\rightarrow$Real} benchmark, where the models are evaluated on  real videos derived from Toyota Smarthome, an existing ADL dataset~\cite{das2019toyota}. 
As our visual model, we adopt two off-the-shelf video
classification architectures, which we train with the \textsc{Sims4Action} data only.
Our results constitute a challenging benchmark for future research and have two-fold conclusions:
(1) video game supervision has potential for ADL recognition  in real settings, as all models surpass the random baseline by a large margin, although (2) The  \textsc{Gaming$\rightarrow$Real} evaluation regime is very challenging as modern recognition algorithms are highly susceptible to domain shifts.
We hope that our dataset detaches ADL recognition from heavy physical data collection cycles and  creates a new avenue for future research focusing on capturing the essence of human behaviour in both, synthetic and real domains.

%% file: sections/related_work.tex
\section{Related Work}

\subsection{Recognizing Activities of Daily Living}
The performance of models recognizing domestic activities  is directly linked to the progress in general video classification, which has undergone a sudden shift from machine learning approaches operating on hand-crafted features, such as Hidden Markov Models or Improved Dense Trajectories~\cite{wang2013action,roitberg2015multimodal} to end-to-end Convolutional Neural Networks (CNNs)~\cite{carreira2017quo}.
Today, spatiotemporal 3D CNNs~\cite{tran2015learning, carreira2017quo, xie2018rethinking} are  considered front-runners in activity classification and are used in a wide range of applications, such as robotics, autonomous driving or healthcare~\cite{wen2019human, roitberg2020cnn, gebert2019end, roitberg2020open,  sharghi2020automatic}.

Research of ADL recognition for domestic robotics or smart homes often follows similar classification frameworks and focuses on constructing annotated datasets with the specific application requirements in mind~\cite{dai2020toyota,das2019toyota,shahroudy2016ntu,wang2018free,Sigurdsson2016HollywoodIH,jang2020etri}.
The creators of the Toyota Smarthome dataset~\cite{das2019toyota,dai2020toyota}, for example, equipped apartments of elderly people with multiple  Kinect cameras and leverage I3D enhanced with spatio-temporal attention to distinguish $31$ activity classes.
While Toyota Smarthome is an excellent benchmark, recording and annotating such data requires heavy effort and extending it with new  behaviours, environments or  appearance types is harder than reproducing desired situations inside rich simulation games such as \textsc{The Sims 4}.
We aim to overcome this labour-intensive constraint and explore ADL recognition by learning from life simulation video games, which, to the best of our knowledge, has not been considered yet.
While our main contribution is the \textsc{Sims4Action} dataset with domestic activities recorded during gameplay, we establish correspondences between our classes and multiple Toyota Smarthome categories in order to validate if such videogame-based training examples can be used to classify real data. 


\subsection{Leveraging Video Games and Simulations for  Data}
Recent works have considered video games as a source for inexpensive training data especially focusing on autonomous driving applications.
Richter \etal~\cite{richter2016playing, richter2017playing} presented a method to collect data and ground truth for visual perception tasks in driving scenarios (semantic segmentation, visual odometry, optical flow, object detection \& tracking) from the video game Grand Theft Auto V (GTA V).
Kr\"ahenb\"uhl~\cite{krahenbuhl2018free}, presented an improved data collection method: fully automated ground truth extraction from video games, without human annotation effort.
Through a wrapper around the DirectX 11 API the rendered data is obtained in real-time, i.e. during actual gameplay.
The authors collected 220 thousand training- and 60 thousand test images across three different games (Far Cry Primal, The Witcher 3, GTA V) and were able to gather rich image meta information such as albedo, depth, instance segmentation, semantic labeling, optical flow and occlusion boundaries.
Our dataset is created with a different type of game - the extensive life simulator \textsc{the Sims 4} and targets human behaviour, while previous work has strong focus on the underlying scene and video features, such as segmentation maps or optical flow.

While learning behaviours from synthetic data is an underresearched area, few recent works have introduced simulation tools explicitly targeting human activity recognition~\cite{varol2021synthetic,puig2018virtualhome,hwang2020eldersim}. 
The research most similar to ours is presumably the concurrent work of Varol \etal~\cite{varol2021synthetic}, which leverage Blender-based simulations to improve activity recognition from viewpoints not present during training.
A different line of work considers activity simulations from the perspective of body poses, synthesizing skeleton trajectories with neural networks~\cite{guo2020action2motion, blattmann2021behavior}.
In contrast to the previous research, we  focus on learning new  categories of daily activities with \emph{videogame supervision},  which, to the best of our knowledge, is explored for the first time.

%% file: sections/dataset.tex

\section{Playing Sims 4 for ADL Training Data}
\begin{figure*}\centering
	\includegraphics[width=\textwidth,trim={0 13cm 0.8cm 0},clip]{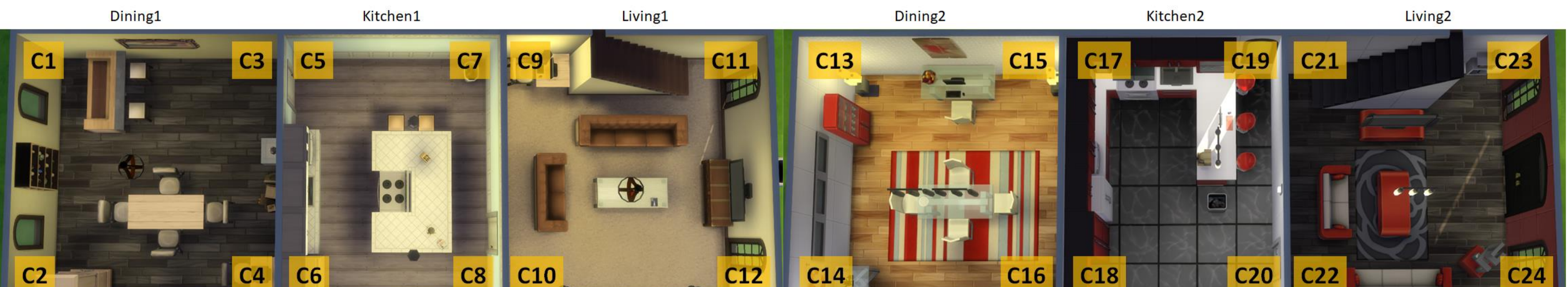}
	\caption{Snapshots of the virtual apartments we designed to simulate household situations.
	The two homes cover a dining room, a living room and a kitchen.
	\textit{C1} - \textit{C24} marks the $24$ camera placements used in the static camera mode recordings.}\hfill
	\label{fig:scenes}
\end{figure*}

\begin{figure}
	\includegraphics[width=0.5\textwidth,height = 3.8cm]{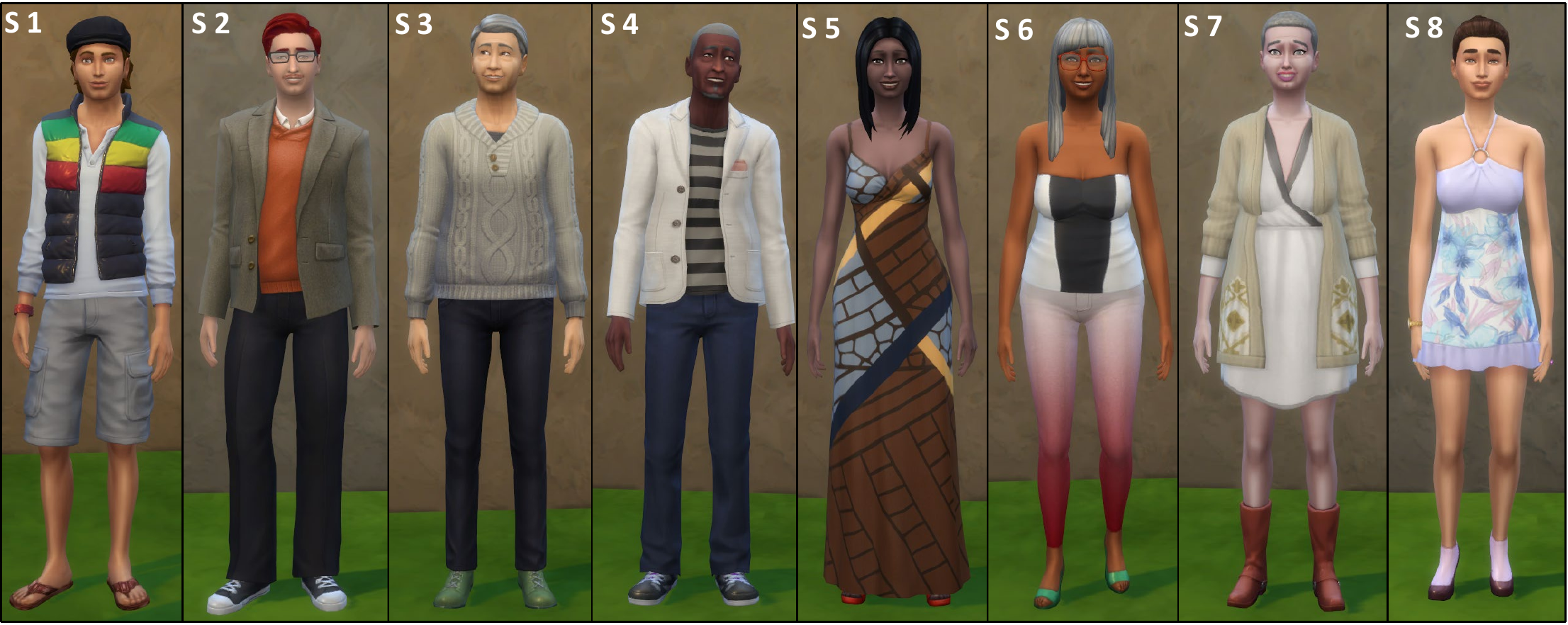}
	\caption{Eight subjects (four elderly people, two adults and two young adults) are covered in the \textsc{Sims4Action} dataset.}
	\label{fig:subjects}
\end{figure}

We introduce \textsc{Sims4Action} -- a dataset designed for visually recognizing human behaviour at home by learning from life simulation video games. 
Our dataset covers  ten different behaviours carried out by eight designed subjects, while the gaming situation allows us to control their appearance and environments according to our needs. 
The main characteristics of our dataset are provided in Table \ref{tbl:stas}.

 


\subsection{Data Collection}



  \textsc{Sims4Action} dataset is collected using \textsc{the Sims 4} life-simulation video game, where players guide the virtual humans through everyday activities of all aspects leading to complex character traits, varying emotional states, diverse subject appearances, and ages.
  We  leverage this environment to induce the situations we want to recognize and use the built-in video capture tool for the recordings (the tool allows us to select the time frame, capture size and image resolution, which is $30$ FPS and $640 \times 368$ pixels in our case).
The main stages of our data collection are now described in  detail and can be summarized in 1) design of the virtual human 2) design of the home environment, and 3) actively "playing" behaviour categories we want to recognize while recording the gameplay.

\mypar{Virtual Subjects}
Recruitment of diverse  subjects is essential to overcome dataset biases, but is often difficult in practice.
Complementing real datasets with samples from life simulation video games might mitigate this issue, as \textsc{The Sims 4} enables detailed subject personalization even capturing details such as walking style. 
Keeping broad range of appearances in mind, we have designed eight virtual humans diversifying their look,  genders (four male and four female) and age groups (four elderly people, two young adults, two adults).
 An overview of the created subjects is provided in Figure \ref{fig:examples_sims_toyota}.

\begin{table}[]
\begin{tabular}{lcccc}
\toprule
\textbf{Property} & \textbf{Train} & \textbf{Validation} & \textbf{Test} & \textbf{Total}\\
\midrule
Duration (minutes) & 388.5 & 79.6 & 157.45 & 625.6\\
Number of samples* & 8217 & 1684 & 3331 & 13232\\
Number of subjects & 5 & 1 & 2 & 8\\
Male/Female & 2/3 & 1/0 & 1/1 & 4/4\\
Static camera & \ding{52} & \ding{52} & \ding{52} & \ding{52}\\
Moving camera &\ding{52}  & \ding{52} &\ding{52} & \ding{52} \\
Number of static views & 24 & 24 & 24 & 24\\
Number of  activity classes & 10 & 10 & 10 & 10\\
Min. nr. samples per class & 769 & 150 & 324 & 1265\\
Avg. nr. samples per class & 821.7 & 168.4 & 333.1 & 1323.2\\
\bottomrule
\end{tabular}
\caption{Characteristics of the \textsc{Sims4Action} dataset.\\
$^\star$A sample is a three second snippet with its assigned label.}
\label{tbl:stas}
\end{table}

\begin{figure}[t]
	\includegraphics[width=0.5\textwidth]{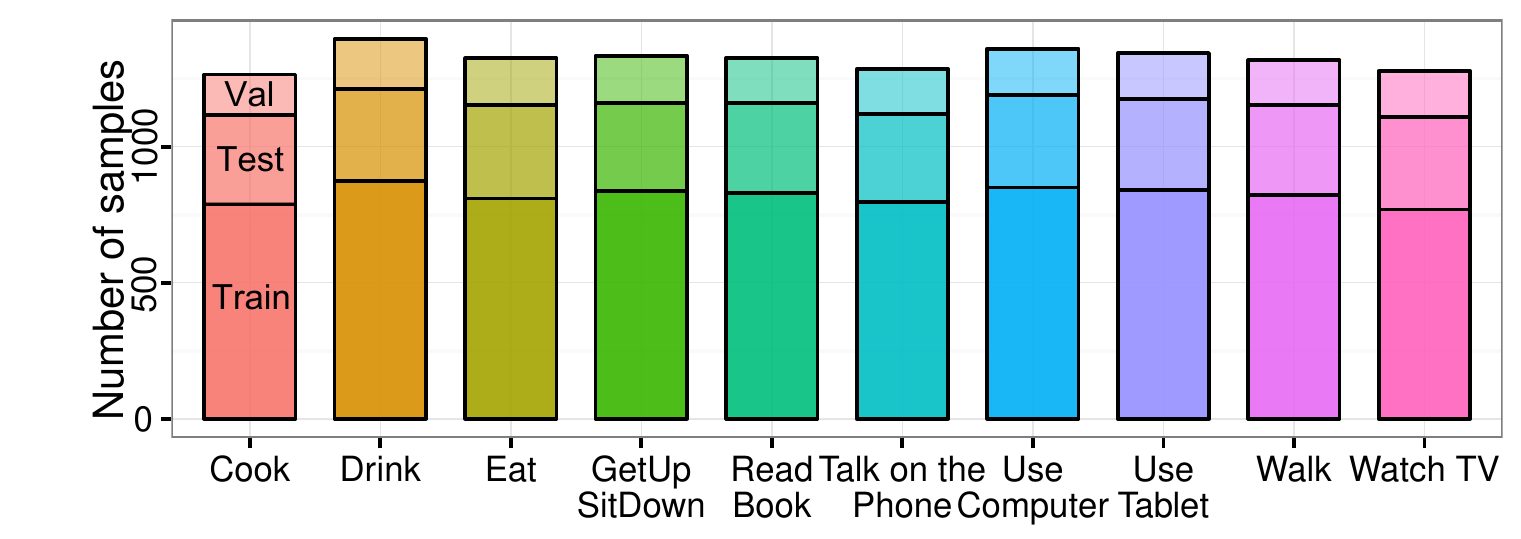}
	\caption{Sample frequency of the activities in  \textit{train}, \textit{test} and \textit{val}. In contrast to real-life data collection~\cite{das2019toyota}, life simulation video games give us control of the data distribution resulting in a well-balanced dataset. (A sample corresponds to a $3$s snippet with the assigned label.)}
	\label{fig:sample_frequency}
\end{figure}

\mypar{Inducing Virtual Behaviours}
In \textsc{the Sims 4}, the virtual subject (Sim) follows given instructions. Thus, we explicitly enforce the behaviors-of-interest in a top-down manner, by repeatedly guiding the Sim to engage in the desired activities, which range from cooking or taking pills to rare events such as having a wedding party or morning sickness during pregnancy. 
In this work, we focus on ten everyday household activities popular in previous works on recognition in real smart homes.
Examples for each of the classes are provided in Figure~\ref{fig:examples_sims_toyota}.
While balanced datasets are rare in real-life~\cite{das2019toyota},  synthetic data collection enables us to easily design equal distribution of classes (ca. one hour footage per category), although the number of videos for each activity  varies as the activities have different durations (\eg, around 15 to 30 seconds for \textit{drink}, and around 30 seconds to one minute for \textit{cook} and \textit{eat}). The duration of some activities can be controlled by the player (\textit{get up/sit down}, \textit{use computer}, \textit{use tablet}, \textit{walk}, \textit{watch TV}), whereas other activities will be completed automatically after a certain amount of time (\textit{cook}, \textit{drink}, \textit{eat}, \textit{read book}, \textit{use phone}).
Figure \ref{fig:sample_frequency} illustrates sample statistics for the individual categories.

\begin{figure}[t]
	\includegraphics[width=0.5\textwidth]{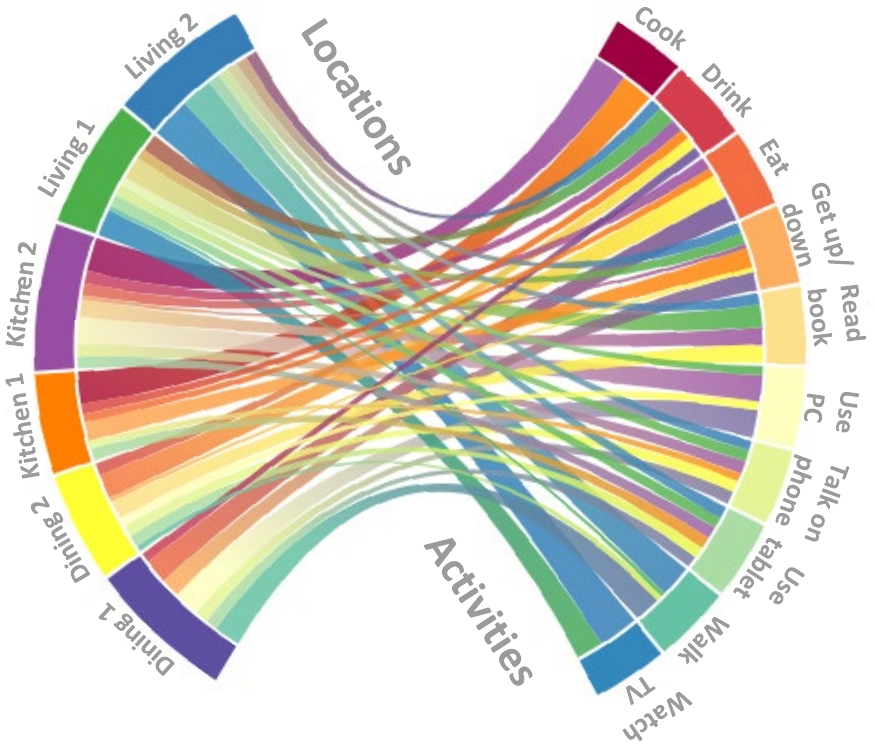}
	\caption{\textit{Where did the activities take place?} The cord diagram visualizes the correspondences of the captured behaviours and the rooms it took place in. Best viewed in color.}
	\label{fig:chord_figure}
\end{figure}

\mypar{Virtual Homes}
As users can create customized interiors, we can generate a large variety of testing grounds.
We chose to lean close to the Toyota Smarthome~\cite{das2019toyota,dai2020toyota} framework and recreate digital versions of their households with the game's build mode.
Our dataset covers two simulated apartments with three rooms: a kitchen, a dining-, and a living room.
While some actvities (such as \textit{talking on the phone}) happen at various locations, other events, such as \textit{cooking} have natural  location biases (the location distribution of the captured activities is provided in Figure \ref{fig:chord_figure}).

The simulation environment allows us to freely  choose the camera angle.
We record the videos with two different cameras: (1) the \textit{fixed camera} mode, where the camera stays in one determined position for the entire video, and (2) the \textit{moving camera} mode, where the camera is set in various random angles and moves for every few seconds in the video.
For the \textit{fixed camera} mode, we define four different camera angles in each room, resulting in  a total of $24$ distinct camera views. 
An overview of the modeled homes and the camera views is shown in Figure~\ref{fig:scenes}.

\subsection{Dataset and Splits Statistics}
\label{sec:splits}
 We collect 942 videos with a total length of 10 hours, 25 minutes, and 32 seconds and a roughly equal representation of activity classes (ca. 1 hour each) and subjects (ca. 1 hour, 17 minutes each).

\mypar{\textsc{Sims4Action} Splits}
The generalization to new humans is essential in ADL recognition.
We therefore divide our dataset \emph{subject-wise}  into \emph{train} (5 subjects), \emph{val} (1 subjects), and \emph{test} (2 subjects). 
We  learn model parameters
on \emph{train}, select checkpoints and hyperparameters on
\emph{val}, and present final evaluation on \emph{test} (overview of the split statistics provided in Figure \ref{fig:sample_frequency}).
Since behaviour complexity and duration vary strongly, we divide the videos into chunks of three seconds and use them as samples in our benchmark.  
Hence, the classification task on our dataset amounts to assigning the correct activity label to a three second video.

\mypar{\textsc{Gaming$\rightarrow$real} Benchmark}
Since neural networks heavily rely on processing the data distribution they were trained on, the task to transfer recognition capabilities from the \textsc{Gaming}$\rightarrow$\textsc{real} domain is challenging.
To enable transfer from video games in actual robotics or smart home applications long-term, \textsc{Sims4Action} is accompanied by a real data benchmark.
To achieve this, we build correspondences between our categories and activities captured in the real Toyota Smarthome dataset~\cite{das2019toyota} and use the Toyota Smarthome cross-subject test set with the corresponding classes for evaluation.
Note, that we do not directly take Toyota Smarthome classes as-is, and certain modifications and consolidations are required to match the \textsc{Sims4Action} classes.
For example, we consolidate different variants of cooking activities into a single cooking class. 
Figure \ref{fig:examples_sims_toyota} compares the videogame examples and their real-life counterparts for every category.
Overall, the \textsc{Gaming$\rightarrow$real} test set comprises $4412$ videos with the corresponding ten activities inside a real smart home, where each video is used as a test sample, as in ~\cite{das2019toyota}.
This real-life test data is not evenly distributed, with the number of samples ranging between $15$ for \textit{use tablet} and $1225$ for the \textit{walk} category.


%% file: sections/approach.tex
\section{Neural Architectures}
\label{sec:approach}

To evaluate our idea of ADL recognition by learning from life simulation video games and provide a competitive benchmark based on the \textsc{Sims4Action} dataset, we adapt two off-the-shelf CNNs for general activity recognition.

\begin{figure*}\centering
	\includegraphics[width=0.96\textwidth,page=1]{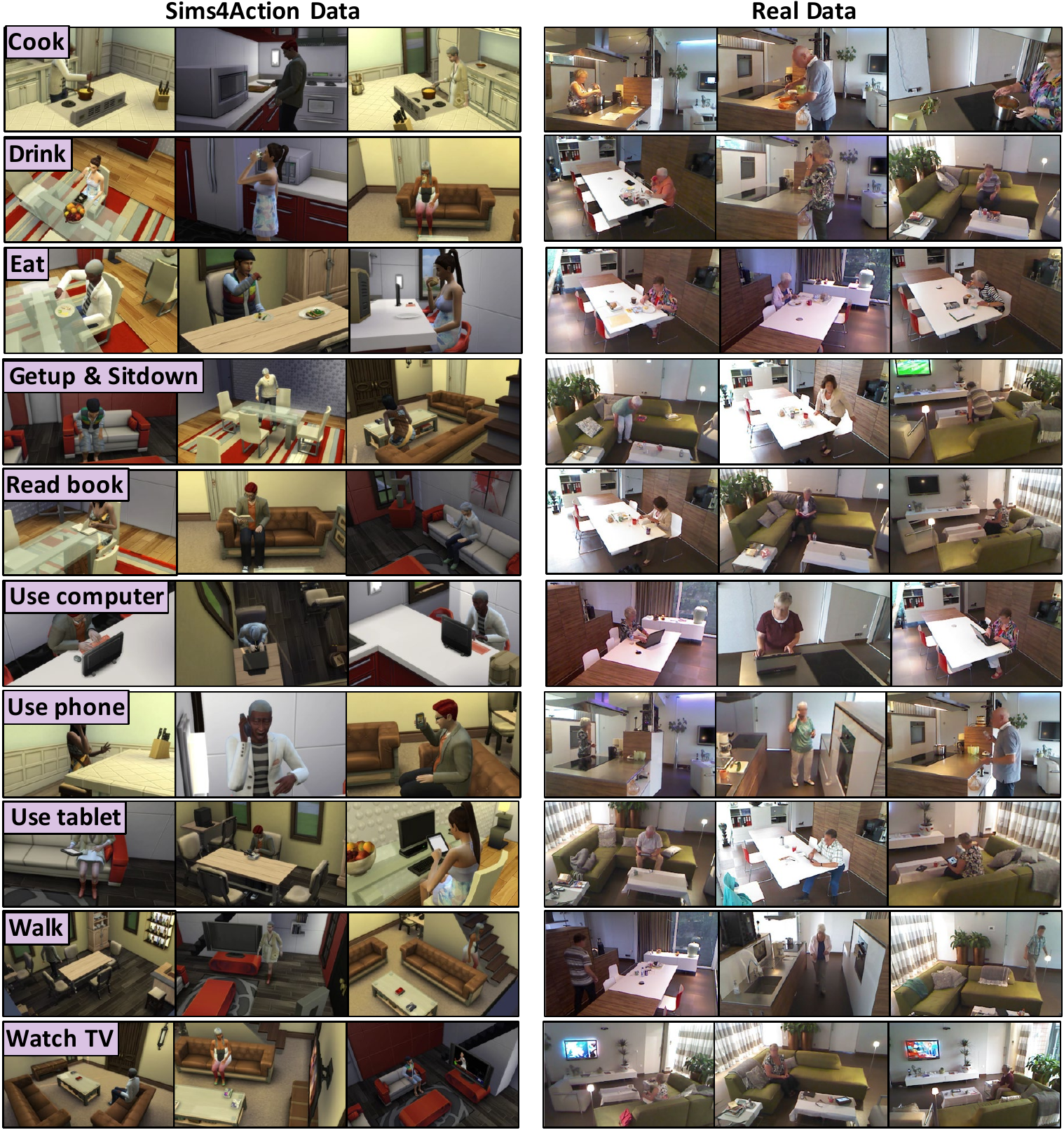}
	\caption{Examples of subjects being engaged into different domestic activities in the \textsc{Sims4Action} dataset and the real-life counterparts used for the \textsc{Gaming$\rightarrow$Real} Benchmark, which are derived from Toyota Smarthome~\cite{das2019toyota}.}
	\label{fig:examples_sims_toyota}
\end{figure*}



\mypar{Inflated 3D CNN}
The Inflated 3D ConvNet (I3D) by Carreira and Zisserman~\cite{carreira2017quo} is a wildly successful video classification architecture utilized for both, general- and ADL recognition~\cite{carreira2017quo,das2019toyota}.
The model benefits from the object recognition architecture Inception-v1~\cite{ioffe2015batch} and modifies it to deal with the temporal domain via expansion of 2D- to 3D kernels and ablations on how to integrate pooling operations.
Inherited by the ancestor network,  I3D is $27$ layers deep and stacks three convolution layers at the beginning, nine inception modules which themselves are two layers deep, one fully-connected layer at the end as well as five max- and average-pooling layers.

\mypar{Separable 3D CNN}
While I3D offers excellent classification accuracy, it is a heavy architecture with $>12$ million parameters and might be impractical in robotic applications. 
With this in mind, Xie~\etal~\cite{xie2018rethinking} redesigned the I3D architecture by considering (1) switching to a top-heavy model, (2) temporally separable convolutions, and (3) spatio-temporal feature gating.
These alterations led to a major reduction in floating point operations per second (FLOPs)  and culminated in the new Separable 3D architecture (S3D).
The so-called top-heavy model property reduces FLOPs by switching early 3D convolutional blocks in the I3D architecture with standard 2D convolutions and only in later layers involve temporally separable 3D convolutions.
The high accuracy of S3D can be attributed to a gating mechanism after temporal convolutions in the separable convolution blocks, which resembles the mechanics of \emph{self-attention}.

\mypar{Implementation and Training Details}
In all experiments except for the baselines, we train our networks end-to-end on \textsc{Sims4Action} for 100 epochs. We use the I3D architecture with an input frame size of $224 \times 224$ or the S3D architecture with an input frame size of $128 \times 128$. The network architecture-related parameters (e.g.dropout, input size) were set according to the original architectures~\cite{carreira2017quo,xie2018rethinking}. 
We apply multiple augmentations during training including random rotations ($\pm$ 20\degree), hue changes ($\pm$ 180\degree), saturation changes ($\pm$ 100\%), brightness changes ($\pm$ 80\%) and random cropping.
For testing, we only scale the input clips to the desired input size and take a center crop.
Conceptually, we divide all videos into 90 frame chunks. During training, we randomly select single $30$ frame clips from each chunk and classify them separately. For the final classification on videos, we only classify the centre chunk of each video.
We use the Adam optimizer with a learning rate of 1e-4 and weight decay 1e-4. For training on the Sims dataset, we used a mini-batch size of 48 for S3D and 20 for I3D.
For our transfer learning experiments on ADL, we exchanged the last layer to match the number of 31 classes and then trained on that layer for one epoch in order to fine-tune the parameters. We used a batch size of 64 chunks for S3D and a batch size of 32 chunks for I3D, all other settings are equal to the Sims training. 

%% file: sections/evaluation.tex

\input{tables/results_sim_to_sim}

\input{tables/prec_recall}

\section{Experiments}

We now investigate how well off-the-shelf activity recognition models described in Section \ref{sec:approach} can leverage our synthetic \textsc{Sim4Action} data to recognize activities of daily living.
To gather this insight, we present results for two scenarios: (1) recognition of ADL in the video game domain and (2) recognition transfer from video game-trained models to real-life ADL classification.
We use the standard accuracy and the balanced multi-class accuracy (average of the individual class recalls) as our main evaluation metrics.

\subsection{ADL Recognition on Video Game Data}

First, we explore how well the described CNN architectures can handle the  video game data in the \textsc{Gaming$\rightarrow$Gaming} setting, where we train and evaluate in the synthetic domain.
In Table~\ref{table:sim_to_sim} we present the results of the S3D and I3D networks both trained on the \textsc{Sims4Action} dataset.
Similarly to the Toyota Smarthome research~\cite{das2019toyota}, we initialize the networks using the Kinetics-400 classification models~\cite{kay2017kinetics}, but also report the results when training from scratch.
As expected, the accuracy of the pretrained models is considerably higher, since 3D CNNs are known to strongly benefit from pretraining.
When granting the models a small margin of error in the \emph{Top 3} evaluation scenarios, they already come close to perfectly recognizing the synthetic ADLs.
In addition to the overall accuracy,  we   examine  model  performance for  the  individual classes, with the exact precision, recall and F1-score (harmonic mean of the precision and recall) provided  in Table~\ref{tbl:prec_recall}.
Interestingly, the best recognition in terms of $F1$-score is achieved for the activities \textit{cook} and \textit{use computer}. This is likely due to the distinctiveness of the cooking utensils and the computer making them easily distinguishable patterns in those activities.
\textit{Get up/sit down} is quite ambiguous and hard to disentangle from other activities like \textit{watch TV} or \textit{eating} where the virtual human is also in a seated position resulting in lower precision.

%
%
%
%

\input{tables/results_sim_to_real}

\subsection{ADL Recognition in the \textsc{Real-world} from Video Games}

Our next area of investigation is the \textit{transfer} scenario, where models trained on \textsc{Sims4Action}  are applied on real videos not present during training.
The complementary real-world benchmark was derived from the Toyota Smarthome dataset~\cite{das2019toyota} (see Figure~\ref{fig:examples_sims_toyota} for images from both datasets), as described in Section~\ref{sec:splits}.
The results of this experiment are reported  in Table~\ref{table:sim_to_real}.
Direct cross-domain  recognition is an important but very hard task for modern data-driven algorithms and the performance also drops  in our case.
Still,  all models clearly outperform the random baseline and $34\%$ of cases were correctly identified by the I3D Net in the case of Kinetics-initialization.
This experiment results provide encouraging evidence, that a cheap, synthetic data collection from sources such as life simulation video games is  a promising direction for training ADL models.
We believe, that such frameworks provide a pathway towards economical data collection in the future.
As our results also demonstrate the sensitivity of modern ADL recognition models to changes in data distribution, we hope that our dataset facilitates development of domain-agnostic models, which is an active topic, \eg, in autonomous driving~\cite{alberti2020idda, ma2021densepass} but has been rather overlooked in ADL recognition.







%% file: tables/results_sim_to_sim.tex
\begin{table}[t]
	\centering

		\scalebox{0.8}{
			\begin{tabular}{@{}llccc|ccc@{}}
			
			\toprule
			\textbf{Model} &\multirow{2}{*}{\makecell{Weight \\ initialization}} & \multicolumn{3}{c|}{\textbf{Normal Acc [$\%$] }} & \multicolumn{3}{c}{\textbf{Balanced Acc [$\%$]}} \\
			&& Top 1 & Top 3 &Top 5 & Top 1 & Top 3 &Top 5\\
			\midrule
			Random Baseline & None & 10.0 & 33.33 & 50.00 &10.0 & 33.33 & 50.00\\
			\midrule
			Separable 3D Net & Random &  56.74 & 79.36 & 91.18 & 56.52 & 79.33 & 91.25\\
			Separable 3D Net & Kinetics400 &84.67 & 96.65 & 98.67 & 84.61 & 96.67 & 98.68\\
			I3D  Net & Random & 67.0 & 87.81 & 94.29 & 66.91 & 87.78 & 94.33\\
			I3D  Net & Kinetics400 & 89.06 & 98.35 & 99.55 & 89.12 & 98.34 & 99.54\\
			\midrule
			\bottomrule
		\end{tabular}
		}
	
	\caption{Evaluation of the ADL recognition in the \textsc{Gaming$\rightarrow$Gaming} setting of our benchmark. The performance metrics are  standard- and balanced accuracies.}
	
	\label{table:sim_to_sim}
	
\end{table}

%% file: tables/prec_recall.tex
\begin{table}[]
	\centering
	\scalebox{1.0}{

    \begin{tabular}{lccc}
    \toprule
\multirow{2}{*}{\textbf{True Activity Class}} & Prec. & Recall & F1 \\
 & {[}\%{]} & {[}\%{]} & {[}\%{]} \\
\midrule
Cook   &  98.99 & 93.61 & 96.22 \\
Drink    & 84.74 & 94.44& 89.33\\
 Eat    & 93.90 & 95.65 &94.77\\
 Getup or Sitdown & 78.44 & 85.37 &81.76\\
Read book    & 97.15 & 88.35 &92.54\\
Use computer    & 99.34 & 94.06 &96.63\\
Talk on phone    & 78.17 & 96.67 &86.44\\
Use tablet    & 96.43  &85.71 &90.76\\
 Walk    &100.00 & 57.37 &72.91\\
 Watch TV     &76.74 &100.00 &86.84\\
\bottomrule
\end{tabular}

	}
		\caption{Per-class performance analysis of  I3D  on the \textsc{Sims4Action} dataset: Precision, Recall and F1 (their harmonic mean) score are calculated for each individual class. 
}
\label{tbl:prec_recall}
\end{table}

%% file: tables/results_sim_to_real.tex
\begin{table}[t]
	\centering

		\scalebox{0.8}{
		\begin{tabular}{@{}llccc|ccc@{}}
			
			\toprule
			\textbf{Model} &\multirow{2}{*}{\makecell{Weight \\ initialization}} & \multicolumn{3}{c|}{\textbf{Normal Acc [$\%$] }} & \multicolumn{3}{c}{\textbf{Balanced Acc [$\%$]}} \\
			&& Top 1 & Top 3 &Top 5 & Top 1 & Top 3 &Top 5\\
			\midrule
			Random Baseline & None & 10.0 & 33.33 & 50.00 &10.0 & 33.33 & 50.00\\
			\midrule
			Separable 3D Net & Random & 19.95 & 45.85 & 64.62 & 12.4 & 34.39 & 58.7 \\
			Separable 3D Net & Kinetics400 &  34.08 & 58.64 & 75.94 & 23.23 & 43.85 & 62.29 \\
			I3D  Net & Random &18.02 & 44.39 & 66.06 & 10.91 & 31.49 & 53.58 \\
			I3D  Net & Kinetics400 & 22.75 & 47.76 & 64.99 & 23.25 & 47.02 & 64.41 \\
			\midrule
			\bottomrule
		\end{tabular}
		}
	
	\caption{Recognition results in the \textsc{Gaming$\rightarrow$Real} setting. 
	Such transfer is a hard task for current recognition models, although all methods surpass the random baseline. }
	
	\label{table:sim_to_real}
	
\end{table}

%% file: sections/conclusion.tex

\section{Conclusion}
In this paper, we present \textsc{Sims4Action}, a  dataset for recognizing activities of daily living by learning from synthetic data recorded from life simulation video games. 
Our dataset covers over $10$ hours of annotated video material with individual video clips set in different simulated environments, subjects of different age groups and ethnicities as well as multiple camera view-points.
We adopt two prominent architectures  for video classification to our task and train them to distinguish ten domestic behaviours with the videogame data only present during training.
To provide an extensive benchmark for comparison, we formalize the \textsc{Gaming$\rightarrow$Gaming}  and \textsc{Gaming$\rightarrow$Real} evaluation settings, which we use to validate the models.
Our dataset will be publicly released upon publication and we believe that \textsc{Sims4Action} has strong  potential to  foster research on not only ADL itself but also on domain adaptation between synthetic and real-world activities and motivate the needed development of generalizable video understanding models.

\mypar{Acknowledgements} This work was supported by the Competence Center Karlsruhe for AI Systems Engineering (CC-KING, \url{www.ai-engineering.eu}) sponsored by the
Ministry of Economic Affairs, Labour and Housing Baden-W{\"u}rttemberg.
This research was also  supported by JuBot project sponsored by the Carl Zeiss Stiftung.